\documentclass[10pt,twocolumn,letterpaper]{article}

\usepackage{cvpr}
\usepackage{times}
\usepackage{epsfig}
\usepackage{graphicx}
\usepackage{amsmath}
\usepackage{amssymb}
\usepackage{mathrsfs}
\usepackage{array}

\usepackage[breaklinks=true,bookmarks=false]{hyperref}

\cvprfinalcopy 

\def\cvprPaperID{****} 

\begin{document}

\title{Learning regression and verification networks for long-term visual tracking}

\author{Yunhua Zhang, Dong Wang, Lijun Wang, Jinqing Qi, Huchuan Lu\\
Dalian University of Technology\\
Dalian, China\\
{\tt\small zhangyunhua@mail.dlut.edu.cn, wdice@dlut.edu.cn, wlj@mail.dlut.edu.cn}\\
{\tt\small \{jinqin, lhchuan\}@dlut.edu.cn}
}

\maketitle

\begin{abstract}
Compared with short-term tracking, the long-term tracking task requires 
determining the tracked object is present or absent, and then estimating 
the accurate bounding box if present or conducting image-wide re-detection 
if absent. 
Until now, few attempts have been done although this task is much closer 
to designing practical tracking systems.
In this work, we propose a novel long-term tracking framework based 
on deep regression and verification networks. 
The offline-trained regression model is designed using the object-aware 
feature fusion and region proposal networks to generate a series of candidates 
and estimate their similarity scores effectively. 
The verification network evaluates these candidates and outputs the optimal one 
as the tracked object with its classification score. It is updated online to adapt 
to the appearance variations based on newly reliable observations. 
The similarity and classification scores are combined to obtain the final confidence 
value, based on which our tracker can determine the absence of the target accurately 
and conduct image-wide re-detection to capture the target successfully when it reappears. 
Extensive experiments show that our tracker achieves the best performance on the
VOT2018 long-term challenge and state-of-the-art results on the OxUvA long-term 
dataset.
\end{abstract}

\vspace{-5mm}
\section{Introduction}
Visual tracking is a fundamental problem in computer vision, which has many practical 
applications including video surveillance, vehicle navigation, to name a few. 
The tracking algorithms can be roughly divided into two branches: short-term tracking 
and long-term tracking. 
For the former one, the tracked object is almost always in the camera's field of view but not 
necessarily fully visible. The tracking algorithm focuses on estimating the accurate positions 
and scales of the target in short-term sequences. 
In recent years, numerous trackers~\cite{ECO,CCOT,MDNet,LSART,DRT,SiameseRPN} have 
achieved very promising results in short-term tracking benchmarks (such as OTB~\cite{OTB2015} 
and VOT2017~\cite{VOT2017}). 
However, the experiments on these benchmarks cannot well evaluate the long-term tracking 
performance of different trackers, and cannot provide valid references for the realistic tracking 
systems. 

\begin{figure}[!t]
\begin{center}
\includegraphics[width=1.0\linewidth]{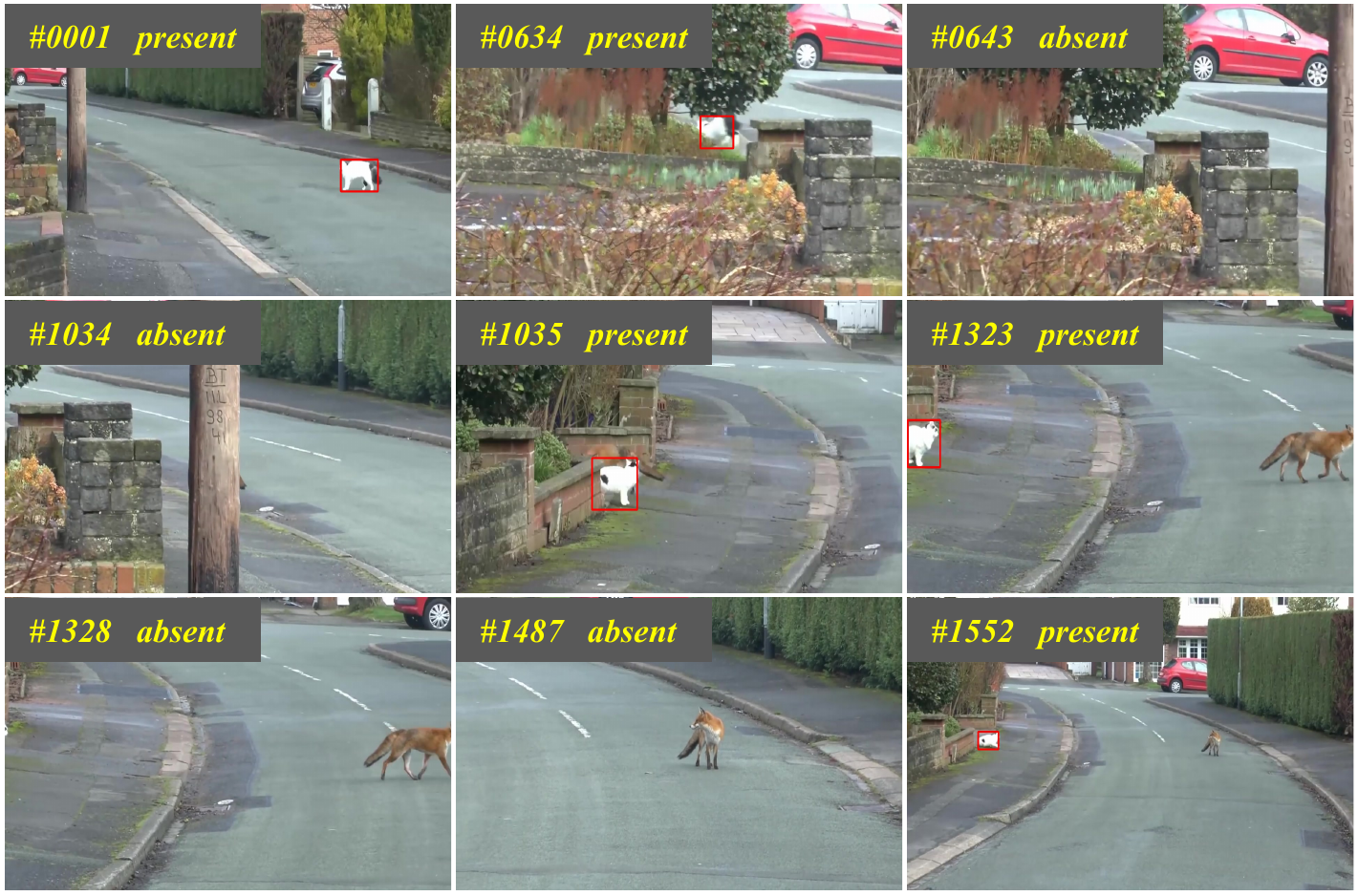}
\caption{Visual results of our tracker in representative frames.}
\label{fig:exp}
\end{center}
\vspace{-5mm}
\end{figure}
\begin{table}[!t]
\caption{Comparisons between recent long-term tracking benchmarks 
and popular short-term ones.}
\vspace{-3mm}
\begin{center}
\scalebox{0.68}{
\begin{tabular}{p{2.5cm}<{\centering}p{1.85cm}<{\centering}p{1.85cm}<{\centering}
p{1.85cm}<{\centering}p{1.85cm}<{\centering}}
\hline 
 & \textbf{LTB35}~\cite{VOTLT} & \textbf{OxUvA}~\cite{OxUvA} 
 & \textbf{OTB100}~\cite{OTB2015} & \textbf{VOT2017}~\cite{VOT2018report}  \\
\hline
Avg frames  &  4196 & 4235 & 590  & 350 \\
Max frames &  29700 & 37440 & 3870  & 1494 \\
Min frames & 1389  &  900 & 72  & 36 \\
absent labels  &  12\% & 52\% & 0\%  & 0\%\\
Avg absent labels & 503 & 2.2 & 0 &  0 \\
\hline
\end{tabular}}
\end{center}
\label{tab:dataset}
\vspace{-8mm}
\end{table}

Recently, the importance of long-term tracking has been emphasized and some large-scale 
datasets have been well constructed to address this issue (such as VOT2018 LTB35~\cite{VOTLT} 
and OxUvA~\cite{OxUvA}). 
Besides common challenges in short-term scenarios, in the long-term tracking task, the tracked object requires to be captured in long-term sequences, and also disappears and reappears very frequently. Thus, this task provides more challenges than short-term tracking. 
Table~\ref{tab:dataset} reports some statistics on the frame length and the number of absent labels in popular and recent short-term (OTB, VOT2017) and long-term (VOT2018 LTB35 and 
OxUvA) benchmarks. First, the frame length in long-term benchmarks is almost ten times 
longer than that in short-term benchmarks. In addition, there exist a large amount of 
\emph{absent} labels in long-term tracking scenarios. 
Thus, it is vital for long-term trackers to provide a valid confidence score indicating the tracked object is \emph{present} or \emph{absent} and have the capability of image-wide re-detection. 
Figure~\ref{fig:exp} provides visual results of our tracker in one representative sequence, 
which shows that our tracker can effectively identify the tracked object is \emph{present} 
or \emph{absent} and estimate the accurate bounding box when it is \emph{present}. \par
Up to now, few works have been done to deal with long-term challenges~\cite{FCLT,LCT,PTAV,MUSTer,CMT}, 
The LCT~\cite{LCT} and PTAV~\cite{PTAV} trackers just track and re-detect the target in a local search region, and cannot re-capture the target successfully after the target moves out of view and reappears again. 
The CMT~\cite{CMT} and FCLT~\cite{FCLT} methods merely exploit a single matching or classification 
model for the entire tracking process, which makes the tracker easily drift due to online variations. 
The MUSTer algorithm~\cite{MUSTer} exploits ensemble models to treat short-term and long-term 
scenarios separately; however, its performance is not satisfactory mainly due to the adopted hand-crafted 
features. 

\begin{figure*}[!t]
\begin{center}
\includegraphics[width=1.0\linewidth,height=0.4\linewidth]{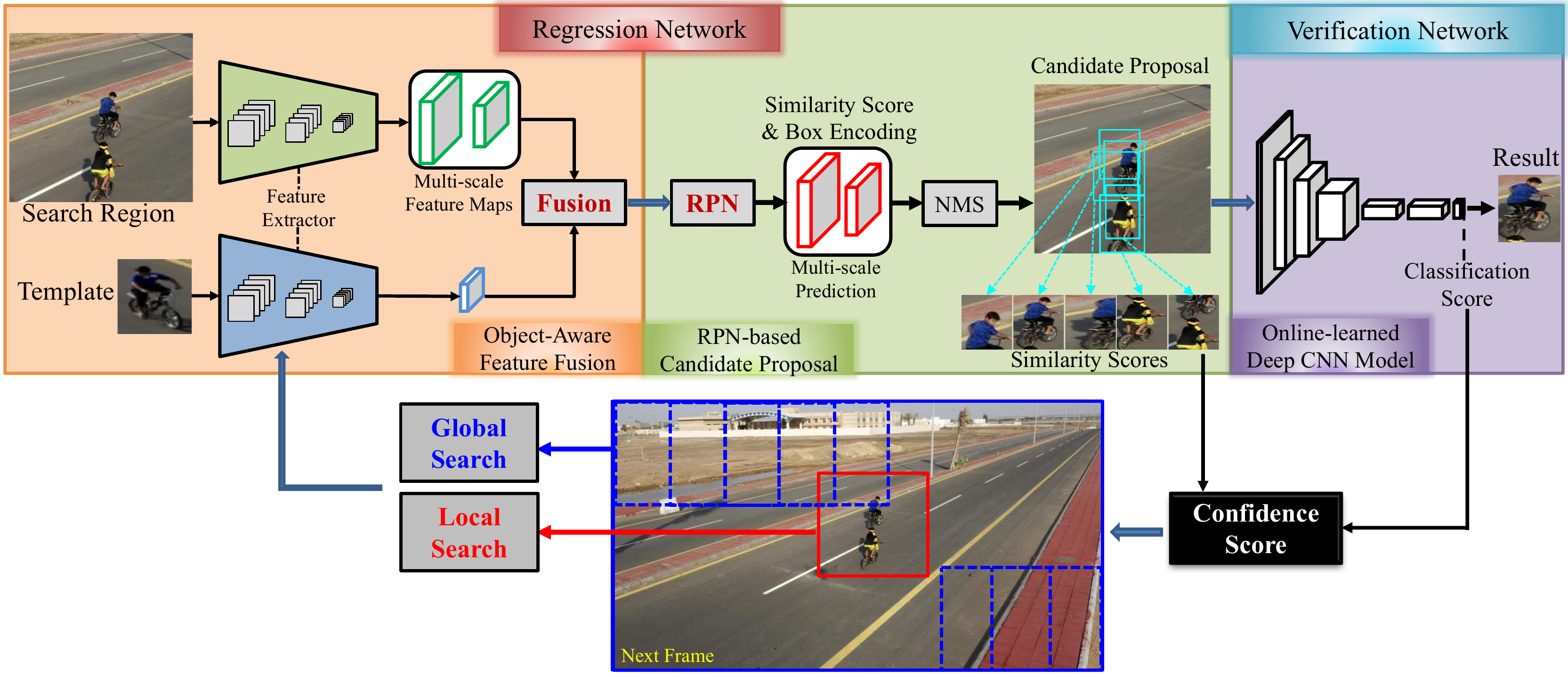}
\end{center}
\caption{The overall framework of the proposed tracking approach. }
\label{fig:framework}
\end{figure*}

To deal with challenges in the long-term scenarios, we attempt to develop a deep-learning-based 
long-term tracker with an integration of regression and verification networks (Figure~\ref{fig:framework}). 
The regression network $\mathcal{R}$ learns a generic matching function off-line to robustly handle 
the common appearance variations of the tracked object during the tracking process. The verification network $\mathcal{V}$ further equips the tracker with a strong discriminative power by online learning. 
In each frame, $\mathcal{R}$ regresses a series of candidate bounding boxes in a local search region, with their scores measuring the similarities between candidates and the object template. 
Then, $\mathcal{V}$ learns a classification boundary online to further decide whether the most 
similar candidate is the true target or a distractor. 
The final confidence score is outputted by the integration of the scores from both $\mathcal{R}$ and 
$\mathcal{V}$, and indicates the tracked object is \emph{present} or \emph{absent} in the current frame. 
This score will be used to invoke the image-wide re-detection scheme if necessary. \par
To summarize, \textbf{the main contributions of this work are presented as follows}: i) A novel long-term tracking framework is developed to combine an offline-learned regression network with an online-updated verification network. The regression model aims to candidate proposal, while the verification one is for target identification. ii) A novel object-specific regression network is proposed to generate a series of candidates being similar with the tracked object, which is offline learned effectively and handles intrinsic appearance variations robustly. iii) A valid confidence score is designed to determine the target is present or not, and to make the tracker dynamically switch between local search and global search. iv) Extensive evaluations on the VOT2018 LTB35 and OxUvA long-term benchmarks demonstrate that our tracker achieves the best performance in comparisons with other competing methods. 
%
%
%
%

\section{Related Work}
\vspace{-2mm}
{\flushleft \textbf{Short-term deep trackers.}}
Recent deep trackers~\cite{MDNet,CCOT,ECO,LSART,DRT,SiameseRPN} have 
achieved promising results in short-term sequences, which are usually categorized into 
either matching-based~\cite{SiameseFC,SINT,SiameseRPN} 
or classification-based~\cite{FCNT,STCT,MDNet,CCOT,ECO,LSART,DRT,DeepSRDCF} ones. 
The former ones attempt to train generalized deep neural networks offline, which find 
the best candidate being most likely to the object template in each frame. 
The classification-based trackers learn discriminative correlation 
filters~\cite{CCOT,ECO,LSART,DRT,DeepSRDCF} or CNN-based classifiers 
online~\cite{FCNT,STCT,MDNet} to distinguish the target from the cluttered 
background. 
The offline-trained matching models are efficient but not well adapt to online variations. 
The online-updated classifiers have powerful discriminative abilities but are 
sensitive to noisy observations. 
Some works~\cite{PTAV,SASiam} have combined both two networks and achieved high 
performance in short-term scenarios; however, they cannot work 
well for long-term tracking~\cite{VOTLT} due to the limitation of their frameworks and re-detection schemes. 
In this work, we develop a novel long-term tracking framework to effectively integrate the 
regression (matching) and verification (classification) networks. 

\vspace{-2mm}
{\flushleft \textbf{Long-term tracking.}}
%
Until now, few works have been proposed for long-term tracking. 
The TLD~\cite{TLD} algorithm exploits the `optical flow'-based tracker for local search 
and an ensemble of weak classifiers for image-wide re-detection. 
The MUSTer~\cite{MUSTer} method utilizes a classifier for short-term localization and 
key points matching for global search. 
In addition, the CMT~\cite{CMT} tracker merely conducts key points matching for long-term tracking. 
The above-mentioned trackers can search the target in the entire image, but their performances are not satisfactory due to the adopted hand-crafted low-level features. 
The LCT~\cite{LCT} and PTAV~\cite{PTAV} trackers are equipped with the re-detection scheme for 
long-term tracking, but they merely track the targets in a local search region to expect that the lost 
targets will reappear around the previous location.  
Thus, these two methods are not able to capture the target any more after it moves out of view. 
FCLT~\cite{FCLT} learns correlation filters online and gradually increases 
the search range with time, but its performance is still far from state-of-the-art. 
This work proposes a novel deep-learning-based long-term tracking algorithm, which 
integrates regression and verification networks effectively. The former one generates a series 
of candidates based on object-aware feature fusion and region proposal 
network (RPN)~\cite{FasterRCNN}\footnote{RPN is a very popular and effective technique 
in object detection~\cite{FasterRCNN,SSD,YOLO9000}. SiameseRPN~\cite{SiameseRPN} 
have attempted to exploit the RPN method for short-term tracking and achieved a highly competitive 
performance. But it needs substantial labeled video data for offline training.}.   
The verification network is an online-learned CNN classier to evaluate the generated candidates. 
It localizes the tracked object accurately and invokes image-wide re-detection if necessary.


\section{Proposed Tracking Approach}
\subsection{Overview}
We propose a long-term tracking framework (Figure~\ref{fig:framework}), with a 
regression network $\mathcal{R}$ regressing a series of bounding boxes that are 
similar to the tracked object and a verification network $\mathcal{V}$ verifying the 
selected candidates.  \par
\vspace{-1.5mm}
{\flushleft \textbf{Candidate Regression.}}
The regression network $\mathcal{R}$ is trained off-line to locate a $127\times127$ 
template image within a larger $300\times300$ search image. 
The search region $x$ and the template $z$ are processed by two different transformations 
$\varphi_1$ and $\varphi_2$, i.e., $\varphi_1(x)$ and $\varphi_2(z)$, respectively. 
A matching function $f(x,z)=m(\varphi_1(x), \varphi_2(z))$ and a regression function 
$g(x,z) = r(\varphi_1(x),\varphi_2(z))$ are learned in a single network to densely compare 
the template image to each candidate region in the search image and regress bounding boxes 
of those are similar to the target. 
The function $m$ is a similarity metric and the function $r$ encodes the location information. 
The most similar ones evaluated by $\mathcal{R}$ are collected to form a candidate pool. \par
\vspace{-1.5mm}
{\flushleft \textbf{Verification.}}
The candidate with the highest similarity score is first cropped out and resized to 
$107\times107$, and verified by the network $\mathcal{V}$. 
This network learns a classification function online to further filter out 
distractions appearing during tracking, where $c_i$ indicates the $i$-th candidate.  
If the most similar one is classified into foreground, the proposed tracker will take 
it as the tracking result $c^*$ of the current frame. 
Otherwise, $\mathcal{V}$ selects a foreground candidate from the candidate pool 
($[c_1, c_2, ... , c_{N_r}]$, $N_r$ is the number of candidates being considered) 
with the higher similarity score than other foregrounds as the tracking result. \par
\vspace{-1.5mm}
{\flushleft \textbf{Re-detection.}}
When both $\mathcal{R}$ and $\mathcal{V}$ cannot find any candidate with high similarity 
and classification scores simultaneously, our tracker regards the tracked object being out of view and searches it in the entire image. 
Unless the tracker has found one patch that is convincing for both $\mathcal{R}$ and 
$\mathcal{V}$, the tracker regards the target as \emph{absent} in the current frame. \par
Since $\mathcal{R}$ is fixed (both the parameters and the target template) during the 
tracking process, it would not accumulate errors and can provide reliable similarity evaluations all the time. 
$\mathcal{V}$ adapts to variations dynamically, and the inaccurate samples collected 
online can be regularized by $\mathcal{R}$ to some extent. 
With the complementation between generalized $\mathcal{R}$ and discriminative 
$\mathcal{V}$, the proposed tracker is capable of searching the target effectively 
in long-term sequences. 
The details are presented in the following sections. 

\subsection{Regression Network}
\label{sec-regression}
The pipeline of our regression network is shown in Figure~\ref{fig:framework}. 
It adopts the SSD~\cite{SSD} detection framework and MobileNets~\cite{MobileNets} 
as feature extractors. 
The two streams of the network share the same architecture. 
Since the sizes of the target are not identical in two inputs (i.e. the search region image 
patch $x$ and the object template $z$), the two branches use different parameters. 
The upper branch takes the search region (cropped around the location of the target 
in the last frame) as input.
It outputs two scales of feature maps $\varphi_1(x)$, namely $19 \times 19 \times 512$ 
and $10 \times 10 \times 512$, to handle drastic scale variations. 
The lower branch takes the object template (ground truth given in the first frame) as input 
and outputs a single $1\times1\times512$ feature vector $\varphi_2(z)$. 
The feature maps of the object template and the search region are then fused and sent 
into the region proposal networks (RPNs). 
The outputs of RPNs are a series of feature maps encoding bounding box information 
and matching results. 
Non-maximum-suppression (NMS) is performed afterwards to get the candidate bounding 
boxes, with threshold of $IoU$ being $0.6$. 

\begin{figure}[!ht]
\begin{center}
\includegraphics[width=1.0\linewidth,height=0.5\linewidth]{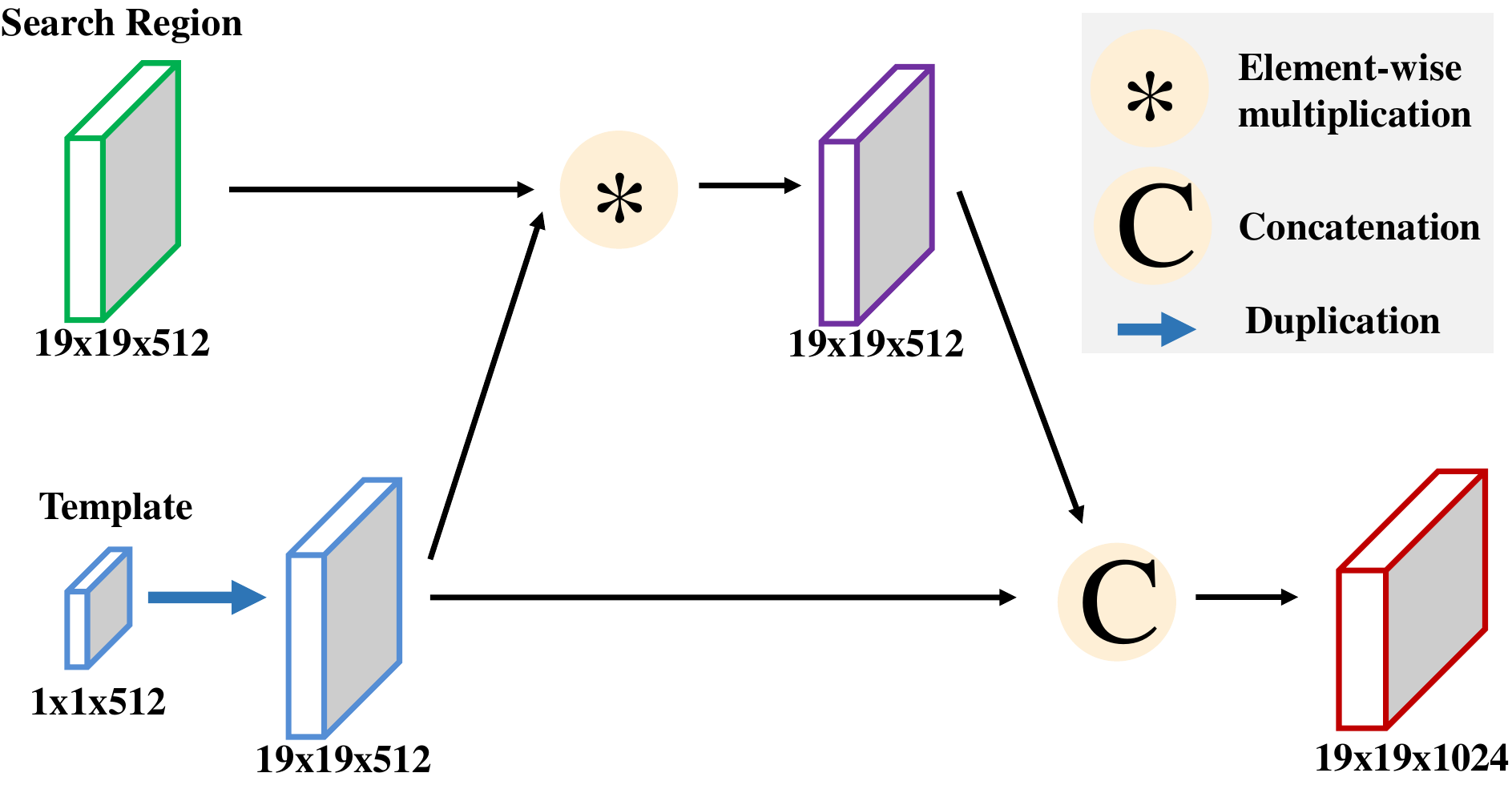}
\end{center}
   \caption{Illustration of the fusion procedure. }
\label{fig:fusion}
\end{figure}

{\flushleft \textbf{Object-Aware Feature Fusion.}}
The fusion procedure aims to provide the region proposal networks with representative 
features of the search region and the tracked object conducting similarity measure and 
bounding box prediction. 
To merge with the feature maps of the search region, the single $1\times1\times512$ 
feature vector of the object template $\varphi_2(z)$ is duplicated to $19\times 19 
\times 512$ and $10\times10\times512$ feature maps. 
The obtained feature maps of the object template have the same sizes with these of the 
search region $\varphi_1(x)$ and are fused with them correspondingly. 
We take the fusion procedure of the $19\times 19$ scale for example, as shown in Figure~\ref{fig:fusion}. 
The $19\times19\times512$ feature maps of the search region are first multiplied with 
these of the object template, which highlights the locations that are similar to provide 
similarity information for the latter metric evaluations. 
The result feature maps are then concatenated with the $19\times19\times512$ feature
maps of the object template to give the latter RPN prior information about the target for 
accurate bounding box regression. 
The final $19\times19\times1024$ feature maps are the inputs of the corresponding RPN. 
The fusion operation of the $10\times10$ scale is the same as that of the $19\times19$. \par

{\flushleft \textbf{RPN-based Candidate Proposal.}}
The proposed tracker has one region proposal network for each scale. 
Each subnetwork has two branches, one for similarity calculation and the other  
for proposal regression. 
The branch for metric learning takes the fused feature maps rather than the 
original feature maps of the image patch as inputs, which is different from 
traditional detection frameworks~\cite{SSD,YOLO9000}. 
Each branch consists of three convolutional layers with $3\times3$ and $
1\times1$ kernels. 
If there are $k$ anchors, for each scale, the network needs to output $2k$ channels 
for matching and $4k$ channels for regression. 
When training the network with several anchors and several scales, we employ 
the loss function that is used in SSD~\cite{SSD}. 
Our loss function is defined as a sum of the matching loss (cross-entropy loss) and 
the localization loss (smooth $L_1$ loss with normalized coordinates). 

\subsection{Verification Network}
In Section~\ref{sec-regression}, the regression network $\mathcal{R}$ generates 
foreground candidates being similar to the object template. 
However, the generated candidate pool may contain distractions that lead to drifts. 
It is inappropriate to directly update $\mathcal{R}$ with respect to a specific object, 
since the errors are inevitably accumulated as the tracking process proceeds.
Thus, a reasonable manner is to keep $\mathcal{R}$ fixed for reliable similarity 
measure and introduce an additional verification network $\mathcal{V}$ for candidate 
evaluation. 
The architecture of $\mathcal{V}$ is similar to that of MDNet~\cite{MDNet}. 
It takes a $107 \times 107$ patch as input and outputs two neurons indicating the 
probabilities of foreground and background respectively. 
More details about the network architecture can be found in~\cite{MDNet}. \par
Similar with the original MDNet, we update the last three convolutional layers of the 
network online to train a strong softmax-based classifier which can distinguish the 
foreground from the background effectively. 
Through online updating, $\mathcal{V}$ helps the proposed tracker tackle with 
various cluttered background during tracking. 
Since the training samples are assessed by both $\mathcal{R}$ and $\mathcal{V}$, 
$\mathcal{V}$ is not likely to break down due to inappropriate updates.   

\subsection{Tracking Strategy}
The proposed tracker first searches the tracked object in the search region, which is four 
times of the object size. 
After obtaining the best candidate in each frame, we can consider the tracked object 
as \emph{present} or \emph{absent} based on its confidence score, and determine 
how to search the target in the next frame. 
If the confidence score ${S_c}$ is below a threshold of $0.3$, the tracker regards 
the tracked object as \emph{absent} and invokes the global search scheme in the 
next frame. 
Otherwise, the tracker treats the tracked object as \emph{present} and continues 
to adopt the local search in the next frame. 
As shown in Figure~\ref{fig:framework}, the local and global search schemes are 
dynamically switched based on the confidence score of the best candidate, which indicates 
whether the tracker finds a reliable candidate or not. \par
In this work, the confidence score ${S_c}$ of the selected candidate in each frame 
is defined by both the regression score ${S_r}$ and the verification score ${S_v}$ as 
\begin{equation}
{S_c} = \left\{ \begin{gathered}
1.0,\;\;{S_v} > {\theta _{{v'}}}\;or\;{S_r} > {\theta _{{r'}}},{S_v} > 0 \hfill \\
0,\;\;\;\;\;{S_r} < {\theta _r},{S_v} < 0 \hfill \\
{S_r},\;\;\;otherwise \hfill \\ 
\end{gathered}  \right., 
\label{eq-confidence}
\end{equation}
where $\theta_{v'} = 20.0$, $\theta_{r'} = 0.5$ and $\theta_r = 0.3$. 
The principles of equation (\ref{eq-confidence}) include: (1) When $\mathcal{V}$ is 
very confident (${S_v} > {\theta _{{v'}}}$) or both $\mathcal{R}$ and $\mathcal{V}$ 
are confident enough  (${S_r} > {\theta _{{r'}}},{S_v} > 0$), our tracker outputs a 
confidence score of $1.0$; 
(2) When both $\mathcal{R}$ and $\mathcal{V}$ give negative feedbacks, the tracker 
returns a confidence score of $0$; 
(3) Otherwise, the tracker sets the confidence score from the regression one ($S_c = S_r$). 

\clearpage
\subsection{Implementation Details}
\vspace{-1.5mm}
{\flushleft \textbf{Network Architectures and Pre-trained Parameters. }}For $\mathcal{R}$, we use the MobileNet architecture~\cite{MobileNets} as the feature extractor for both branches. 
The architecture of $\mathcal{V}$ is VGGM as in~\cite{MDNet}. 
Both VGGM of $\mathcal{V}$ and MobileNets of $\mathcal{R}$ load parameters pretrained on ImageNet classification task~\cite{ImageNet}, while only the regression network $\mathcal{R}$ is further trained off-line using external datasets. \par
\vspace{-1.5mm}
{\flushleft \textbf{Training Data Preparation. }}
During the training phase of $\mathcal{R}$, sampled pairs are selected from both ILSVRC~\cite{ILSVRC} 
image and video object localization datasets with a random interval. 
For the former dataset, we train $\mathcal{R}$ to make it have the capability 
to regress any kind of object for a given object template. 
To be specific, we choose an object of interest from an image randomly and crop the 
corresponding detection region around the chosen object. 
For the video object localization dataset, the similarity calculation branch of the regression network $\mathcal{R}$ further learns a generic matching function for tracking to tolerate the 
common appearance variations. 
The regression network is trained in an end-to-end manner using the
stochastic gradient descent. 
Because of the need of training the regression branch, some data augmentations are adopted including affine transformation and random erasing~\cite{RandomErasing}. \par
\vspace{-1.5mm}
{\flushleft \textbf{Hyper Parameters and Training Strategy. }}
In long-term sequences, since the target moves out of view frequently and its size often changes dramatically when it re-appears, we adopt two scales with different ratios of anchor.
The anchor ratios we adopt are $[0.33, 0.5, 1, 2, 3]$. 
The strategy to pick positive and negative training samples is also important in our proposed framework. 
The criterion used in object detection task is adopted here that we use $IoU$ together with two thresholds $th_{hi}$ and $th_{lo}$ as the measurement. 
Positive samples are defined as the anchors which have $IoU > th_{hi}$ with their corresponding ground truth. 
Negative ones are defined as the anchors which satisfy $IoU < th_{lo}$. 
We set $th_{lo}$ to $0.5$ and $th_{hi}$ to $0.7$. 
Similarly to the training process of SSD~\cite{SSD}, instead of using all the negative examples, we sort them using the confidence loss of each default box and pick the top ones so that the ratio between the negatives and positives is at most $3:1$. 
There are totally $500,000$ iterations performed during training phase and the batch size is $32$. 
Each batch consists of $16$ pairs from image object localization dataset and $16$ pairs from video object localization dataset. 
We use the $10^{-2}$ and $10^{-3}$ learning rates both for $200,000$ iterations, then continue training for $100,000$ iterations with $10^{-4}$. \par
\vspace{-1.5mm}
{\flushleft \textbf{Online Tracking. }}
During the inference phase, the parameters of the regression network are fixed. 
The object template for matching is the groundtruth given in the first frame. 
We only update the verification network in a similar way to that of the original MDNet~\cite{MDNet}. 

\section{Experiments}
Our tracker is implemented using Tensorflow~\cite{Tensorflow} on a PC machine 
with an Inter i7 CPU (32G RAM) and a NVIDIA TITAN X GPU (12G memory). 
The average tracking speed is $2.7$fps. \emph{We will make our source codes 
(both training and testing codes) be available to the public}. 

We evaluate the proposed method on the VOT-2018 LTB35 dataset~\cite{VOTLT} 
and OxUvA Long-term dataset~\cite{OxUvA}. The detailed comparisons are 
presented as follows. 

\subsection{State-of-the-art Comparisons on VOT-2018 LTB35 Dataset}
The VOT-2018 LTB35 dataset~\cite{VOTLT} is presented in Visual Object Tracking 
(VOT) challenge 2018 for evaluating long-term trackers, which includes $35$ challenging 
sequences of various objects (e.g., persons, car, motorcycles, bicycles and animals) 
with a total frame length of $146847$ frames. Each contains on average 
12 long-term target disappearances, and lasts on average 40 frames. Therefore, this 
dataset can fully evaluates trackers' long-term tracking capabilities, which are essential 
for tracking in the wild. 

Following the evaluation protocol of VOT-2018~\cite{VOT2018report}, the tracking precision 
(\textbf{Pr}), recall (\textbf{Re}) and \textbf{F-score} metrics are utilized for accuracy evaluation. 
Based on the precision and recall, the threshold F-measure $F(\tau_{\theta})$ is defined as 
\begin{equation}
\label{Fscore}
F(\tau_{\theta}) = 2Pr(\tau_{\theta})Re(\tau_{\theta}) / (Pr(\tau_{\theta}) + Re(\tau_{\theta})), 
\end{equation}
where $\tau_{\theta}$ is the threshold. 
Then, the F-score is defined as the highest score on the F-measure plot over all thresholds 
$\tau_{\theta}$ (i.e., taken at the tracker-specific optimal threshold). This manner avoids 
arbitrary manual-set thresholds and then encourages fair evaluation. 
In VOT-2018 LTB35~\cite{VOTLT}, the \textbf{F-score} is the primary long-term 
tracking measure and is used for ranking different trackers. 
In addition, the results of the re-detection experiment are adopted to compare different 
trackers, including the average number of frames required for re-detection (\textbf{Frames}) 
and the percentage of sequences with successful re-detection (\textbf{Success}).

The detailed comparisons are reported in Table~\ref{tab:tab1}. 
From Table~\ref{tab:tab1}, we can conclude that our tracker achieves the top-ranked 
performance in terms of \textbf{F-score}, \textbf{Pr} and \textbf{Re} criteria while 
maintaining a high re-detection success rate\footnote{The \textbf{Re} values of DaSiam\_LT 
and Ours are 0.588216 and 0.587921, respectively. Thus, we mark Ours by red and 
DaSiam\_LT by blue.}.
Especially, the proposed method obtains the best performance among all compared 
trackers in terms of the \textbf{F-score} measure, which is the most important metric 
on the VOT-2018 LTB35 dataset~\cite{VOTLT}. 
The DaSiam\_LT method achieves a slightly worse \textbf{F-score} than our 
tracker ($0.607$ for DaSiam\_LT and $0.610$ for ours), however, 
the success rate of DaSiam\_LT is almost zero, which means the DaSiam\_LT 
tracker fails in the re-detection experiment. 
In contrast, the proposed method achieves a success rate of $100\%$.
We also note that our tracker merely uses the ImageNet Detection~\cite{ILSVRC} 
and ILSVRC~\cite{ILSVRC} video datasets for training, while the DaSiam\_LT method 
adopts much more data for training including ImageNet Detection~\cite{ILSVRC}, 
ILSVRC~\cite{ILSVRC}, Youtube-BB~\cite{Youtube} and 
COCO Detection~\cite{COCO}. 

\begin{figure*}[!ht]
\begin{center}
\includegraphics[width=1.0\linewidth,height=0.93\linewidth]{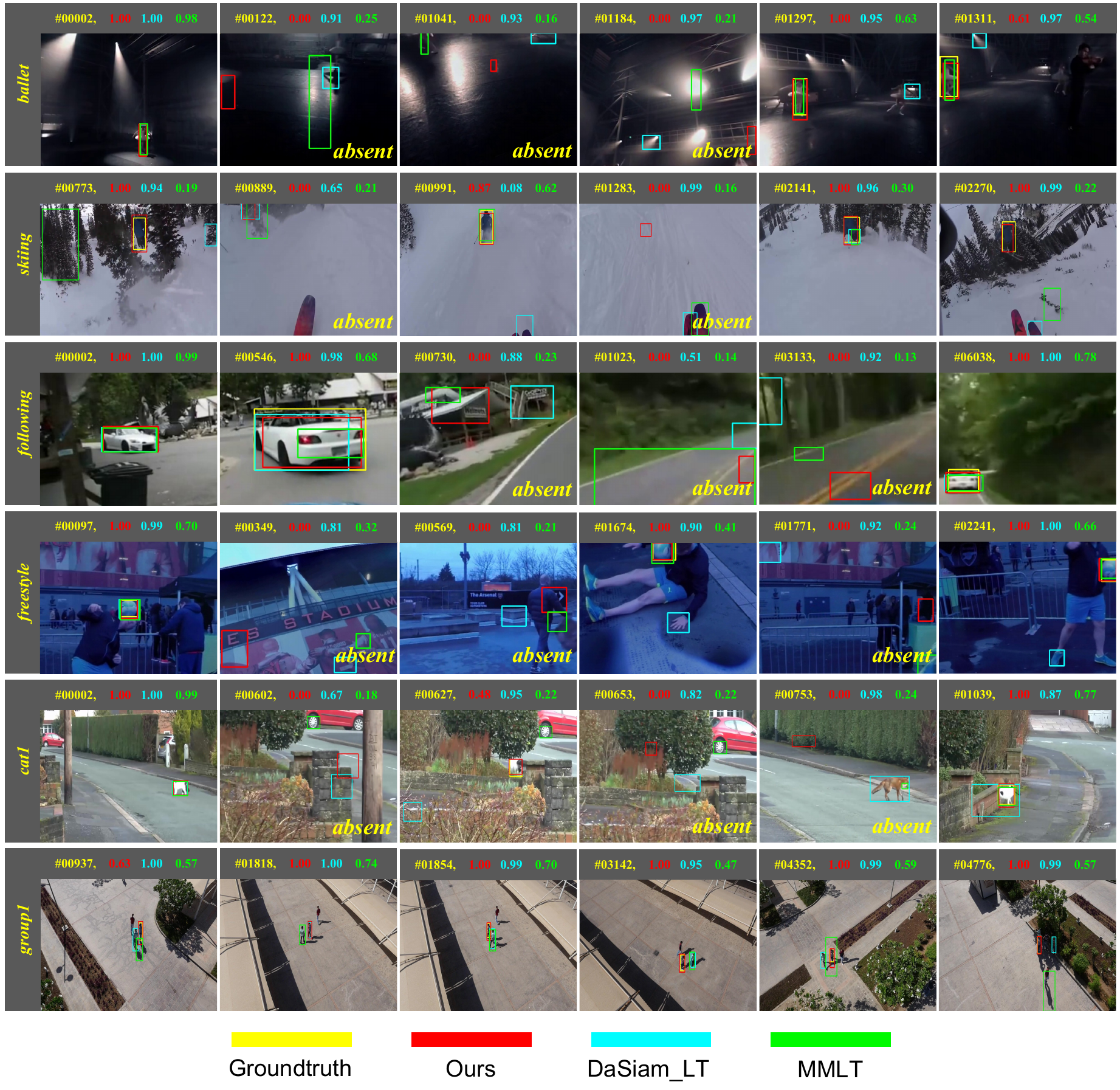}
\end{center}
\vspace{-3mm}
\caption{Qualitative results of our tracker,  along with DaSiam\_LT and MMLT 
(the top-ranked trackers reported in~\cite{VOT2018report}) on six challenging sequences. 
From top to bottom: \emph{ballet}, \emph{skiing}, \emph{following}, \emph{freestyle}, 
\emph{cat1}, \emph{group1}. The confidence score of each tracker is shown in the top 
of each image. We use ``absent" to denote frames in which the target is absent. Best viewed in color with zoom-in for more details. }
\label{fig:qualitative}
\end{figure*}

To visualize the superiority of our tracker, we provide representative results 
of our tracker and other two top-ranked methods reported in~\cite{VOT2018report}, 
along with their confidence scores on challenging image sequences. 
As shown in Figure~\ref{fig:qualitative}, the targets disappear frequently in these videos. 
Our tracker outputs a very low confidence score when the target is absent, while other 
trackers still give relatively higher confidence scores and drift to distractions 
(such as \emph{ballet}, \emph{skiing}, \emph{following}, \emph{freestyle} and \emph{cat1}).
In sequence \emph{group1}, our method captures the tracked object successfully 
but the other two trackers drift to other people easily. 
Thus, the confidence scores outputted by our tracker is able to indicate the presence of 
the tracked object accurately. 
\begin{table}
\caption{Performance evaluation for $15$ state-of-the-art algorithms on the VOT-2018 
LTB35 dataset~\cite{VOTLT}. The best three results are marked in \textcolor{red}{\textbf{red}}, 
\textcolor{blue}{\textbf{blue}} and \textcolor{green}{\textbf{green}} bold fonts respectively.  
The trackers are ranked from top to bottom using the \textbf{F-score} measure.}
\vspace{-5mm}
\begin{center}
\begin{tabular}{p{1.2cm}<{\centering}p{1.2cm}<{\centering}p{0.7cm}<{\centering}
p{0.7cm}<{\centering}p{2.6cm}<{\centering}}
\hline 
\textbf{Tracker} & \textbf{F-score} & \textbf{Pr} & \textbf{Re} & \textbf{Frames (Success)}  \\
\hline
\textbf{Ours}  &  \textbf{\textcolor[rgb]{1,0,0}{0.610}} & \textbf{\textcolor[rgb]{0,0,1}{0.634}} 
& \textbf{\textcolor[rgb]{1,0,0}{0.588}}  & 1 (100\%) \\
DaSiam\_LT  &  \textbf{\textcolor[rgb]{0,0,1}{0.607}} & 
\textbf{\textcolor[rgb]{0,1,0}{0.627}} & \textbf{\textcolor[rgb]{0,0,1}{0.588}}  & - (0\%)\\
MMLT &  \textbf{\textcolor[rgb]{0,1,0}{0.546}} & 0.574 & 
\textbf{\textcolor[rgb]{0,1,0}{0.521}}  & 0 (100\%) \\
LTSINT & 0.536  &  0.566 & 0.510  & 2 (100\%) \\
SYT & 0.509 & 0.520 & 0.499 &  0 (43\%) \\
PTAVplus & 0.481 & 0.595 & 0.404  & 0 (11\%) \\
FuCoLoT &  0.480  & 0.539 & 0.432 & 78 (97\%) \\
SiamVGG & 0.459 & 0.552 & 0.393 & - (0\%) \\
SLT & 0.456 & 0.502 & 0.417 &  0 (100\%) \\
SiamFC & 0.433 & \textbf{\textcolor[rgb]{1,0,0}{0.636}} & 0.328 &  - (0\%) \\
SiamFCDet & 0.401 & 0.488 & 0.341 &  0 (83\%) \\
HMMTxD & 0.335 & 0.330 & 0.339 &  3 (91\%) \\
SAPKLTF & 0.323 & 0.348 & 0.300 &  - (0\%) \\
ASMS & 0.306 & 0.373 & 0.259 &  - (0\%) \\
FoT & 0.119 & 0.298 & 0.074 &  - (6\%) \\
\hline
\end{tabular}
\end{center}
\label{tab:tab1}
\end{table}

\begin{figure}[!ht]
\begin{center}
\includegraphics[width=1.0\linewidth,height=0.9\linewidth]{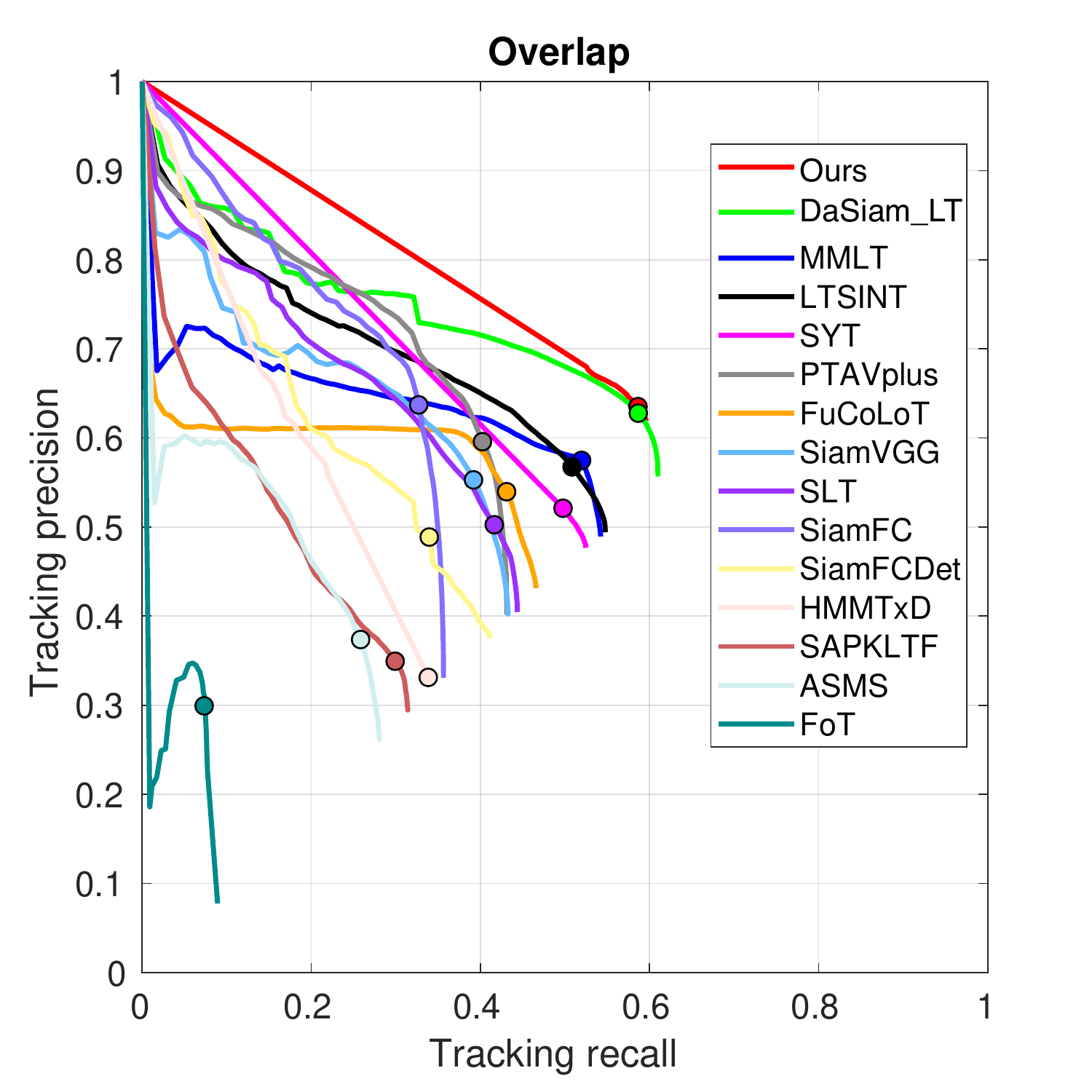}
\end{center}
\vspace{-3mm}
\caption{Average precision-recall curves of our trackers and other state-of-the-art methods 
on VOT-2018 LTB35~\cite{VOTLT}. Different trackers are ranked based on the maximum 
of the F-Score.}
\label{fig:VOT2018}
\end{figure}

\subsection{Ablation Study}
In this subsection, we conduct ablation analysis to evaluate different components 
of our tracker. With different experiment settings, we design four variants of our 
tracker, which are respectively named as ``Ours w/o Verification", 
``Ours w/o Concatenation", ``Ours w/o Multiplication" and ``Ours (Siamese)". 
The meanings of these notions are explained as follows.
(1) ``Ours w/o Verification" denotes the tracker that only uses the regression network 
for tracking, which means that the tracked object is localized as the bounding box 
with the highest confidence score predicted by the regression network in each frame. 
(2) ``Ours w/o Concatenation" and ``Ours w/o Multiplication" stand for different 
operations used in the fusion procedure of the regression network. 
The former one represents the tracker that only uses the result of multiplication 
between the feature of the search region and that of the template without concatenating the 
feature of the template. ``Ours w/o Multiplication" denotes the tracker that only concatenates the feature of the search region with that of the template without multiplication. 
(3) ``Ours (Siamese)" indicates the tracker in which the feature extractors for the 
object template and search region share the same parameters. 

\begin{table}[!h]
\caption{Ablation analysis of the proposed tracker on the VOT-2018 LTB35 dataset. 
The best results are marked in the \textcolor{red}{\textbf{red}} font. }
\vspace{-1mm}
\begin{center}
\begin{tabular}{p{3.8cm}<{\centering}p{1.2cm}<{\centering}
p{0.7cm}<{\centering}p{0.7cm}}
\hline
\textbf{Tracker} & \textbf{F-score} & \textbf{Pr} & \textbf{Re} \\
\hline
Ours  & \textbf{\textcolor[rgb]{1,0,0}{0.610}} 
& \textbf{\textcolor[rgb]{1,0,0}{0.634}} & \textbf{\textcolor[rgb]{1,0,0}{0.588}}\\
Ours w/o Verification & 0.525 & 0.563 & 0.493\\
Ours w/o Concatenation & 0.582 & 0.630 & 0.540\\
Ours w/o Multiplication & 0.442 & 0.504 & 0.394\\
Ours (Siamese) & 0.497 & 0.533 & 0.486\\
\hline
\end{tabular}
\end{center}
\label{tab:ablation}
\vspace{-3mm}
\end{table}

The results of different variants on VOT-2018 LTB35 ~\cite{VOTLT} are presented in 
Table~\ref{tab:ablation}, from which we can obtain the following conclusions. 
(1) ``Ours w/o Verification"  achieves an \textbf{F-score} of $0.525$, which is also 
competitive compared with other state-of-the-art methods (shown in Table~\ref{tab:tab1}). 
It means that our regression network can provide not bad metric evaluation. 
In addition, comparing ``Ours" with ``Ours w/o Verification", we can conclude that 
the proposed verification model greatly improves the performance of our tracker 
for long-term scenario. 
(2) Comparing ``Ours" with ``Ours w/o Concatenation" and ``Ours w/o Multiplication", 
we can conclude that both concatenation and multiplication operations are essential 
in the feature fusion module. 
(3) The comparison between ``Ours" and ``Ours (Siamese)" shows that the Siamese architecture 
(i.e., the feature extractors of the template and the search region share the same parameters) 
leads to inferior performance, mainly caused by different input sizes of these two branches.

\subsection{State-of-the-art Comparisons on OxUvA Long-term Dataset}

\begin{table}
\caption{Detailed comparisons of different trackers on the OxUvA~\cite{OxUvA} 
dataset. The best three results are marked in \textcolor{red}{\textbf{red}}, 
\textcolor{blue}{\textbf{blue}} and \textcolor{green}{\textbf{green}} bold fonts 
respectively.  The trackers are ranked from top to bottom using the \textbf{MaxGM} 
measure.}
\vspace{-1mm}
\begin{center}
\scalebox{1.1}{
\begin{tabular}{p{2cm}<{\centering}p{1.2cm}<{\centering}p{1.2cm}
<{\centering}p{1.2cm}<{\centering}}
\hline
\textbf{Tracker} & \textbf{MaxGM} &\textbf{TPR} & \textbf{TNR}  \\
\hline
\textbf{Ours}  &\textbf{\textcolor[rgb]{1,0,0}{0.544}}& 
\textbf{\textcolor[rgb]{1,0,0}{0.609}} & \textbf{\textcolor[rgb]{0,1,0}{0.485}}  \\
SiamFC+R & \textbf{\textcolor[rgb]{0,0,1}{0.454}} 
& \textbf{\textcolor[rgb]{0,1,0}{0.427}} &0.481 \\
TLD & \textbf{\textcolor[rgb]{0,1,0}{0.431}} & 0.208 
& \textbf{\textcolor[rgb]{1,0,0}{0.895}} \\
LCT & 0.396 &0.292 & \textbf{\textcolor[rgb]{0,0,1}{0.537}} \\
MDNet & 0.343&\textbf{\textcolor[rgb]{0,0,1}{0.472}} &0  \\
SINT & 0.326&0.426 &0  \\
ECO-HC & 0.314&0.395 & 0  \\
SiamFC & 0.313 &0.391 &0 \\
EBT & 0.283&0.321 &0\\
BACF & 0.281& 0.316&0 \\
Staple & 0.261& 0.273&0 \\
\hline
\end{tabular}}
\end{center}
\label{tab:OxUvA}
\vspace{-3mm}
\end{table}

The OxUvA long-term dataset~\cite{OxUvA} comprises $366$ object tracks in $337$ 
videos selected from YoutubeBB~\cite{Youtube} dataset. 
The tracklets are labelled at a frequency of $1$Hz which is the same as YoutubeBB~\cite{Youtube}. 
In this dataset, each video lasts for average $2.4$ minutes, seven times longer than 
OTB-100~\cite{OTB2015} (the most popular short-term tracking dataset). 
Besides the challenging factors in short-term tracking, severe out-of-view and full occlusion 
introduce extra challenges which are close to practitioners' needs. 
The OxUvA dataset is split into \emph{dev} and \emph{test} sets with $200$ 
and $166$ tracks respectively. Using these subsets, two challenges have been defined: 
constrained and open. For constrained challenge, trackers can be developed using only 
data from OxUvA \emph{dev} set (long-term videos). 
For the open challenge, trackers can use any public dataset for training except for the YoutubeBB \emph{validation} set, from which OxUvA is constructed. 
Since all the trackers reported in OxUvA~\cite{OxUvA} are tested for the open 
challenge, we also conduct experiment on the open challenge for fair comparison. 
Following the evaluation benchmark of OxUvA~\cite{OxUvA}, we adopt the true 
positive rate (\textbf{TPR}), true negative rate (\textbf{TNR}), and maximum 
geometric mean (\textbf{MaxGM}) to compare different trackers. 
\textbf{TPR} measures the fraction of \emph{present} objects that are reported as
\emph{present}, and \textbf{TNR} gives the fraction of \emph{absent} objects 
that are reported as \emph{absent}. 
The \textbf{MaxGM} is defined as, 
\begin{equation}
\label{MaxGM}
\begin{array}{l}
\mathbf{MaxGM} \\ 
= \mathop {\max }\limits_{0 \le p \le 1} \sqrt {((1 - p) \cdot \mathbf{TPR})
((1 - p) \cdot \mathbf{TNR} + p)}\\ 
\end{array}, 
\end{equation}
which provides a more informative comparison of different trackers and 
is used to rank them. A larger \textbf{MaxGM} value means a better performance. 
\begin{figure}[!ht]
\begin{center}
\includegraphics[width=1.0\linewidth,height=0.75\linewidth]{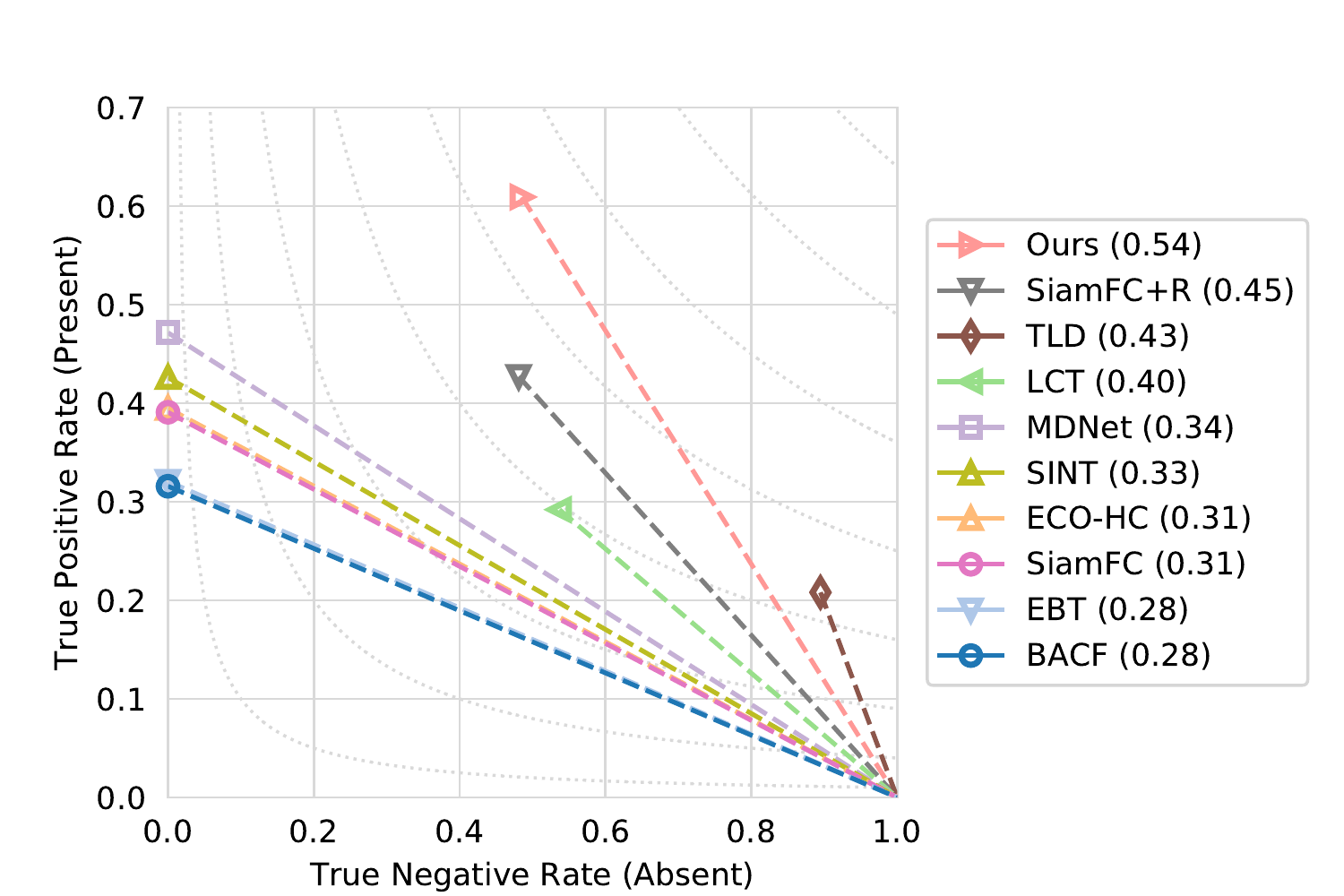}
\end{center}
\vspace{-3mm}
\caption{Accuracy plots of different trackers on OxUvA~\cite{OxUvA}.}
\label{fig:OxUvA}
\end{figure}

We compare our tracker with ten competing algorithms reported in~\cite{OxUvA}, 
including LCT~\cite{LCT}, EBT~\cite{EBT}, TLD~\cite{TLD}, 
ECO-HC~\cite{ECO}, BACF~\cite{BACF}, Staple~\cite{Staple}, 
MDNet~\cite{MDNet}, SINT~\cite{SINT}, SiamFC~\cite{SiameseFC} 
and SiamFC+R~\cite{OxUvA}. 
Among these trackers, LCT~\cite{LCT}, EBT~\cite{EBT} and TLD~\cite{TLD} 
have different re-detection schemes for long-term tracking. 
ECO-HC~\cite{ECO}, BACF~\cite{BACF} and Staple~\cite{Staple} are three 
recent short-term correlation filter trackers with high accuracy and efficiency. 
MDNet~\cite{MDNet}, SINT~\cite{SINT} and SiamFC~\cite{SiameseFC} 
are three popular algorithms based on deep convolutional networks. 
SiamFC+R~\cite{OxUvA} is implemented by equipping SiamFC with a 
sample re-detection scheme similar to~\cite{ODMT}.
If the maximum score of the SiamFC's response is smaller than a threshold, 
the tracker enters the object absent mode. 
In this situation, SiamFC+R attempts to find the tracked object within a random 
search region in each frame until the maximum score again surpasses the threshold, 
at which point the tracker returns to the object present mode. 

The detailed comparisons are reported in Figure~\ref{fig:OxUvA} and 
Table~\ref{tab:OxUvA}. We can see that our tracker achieves the top-ranked 
performance in terms of \textbf{MaxGM} and  \textbf{TPR} while maintaining 
a competitive \textbf{TNR}. 
Especially, our tracker has the best performance among all compared trackers 
in terms of \textbf{MaxGM}, which is the most important metric on the OxUvA 
dataset. 
Compared with the second best method (SiamFC+R), our tracker achieves a 
substantial improvement, with a relative gain of $19.82\%$ in \textbf{MaxGM}.

\section{Conclusion}
In this paper, we propose a novel long-term tracking framework based on regression and verification networks. 
The regression network is trained offline and fixed online for 
effective candidate proposal and reliable similarity estimation 
without accumulating errors in long sequences. 
In addition, the verification network is exploited to evaluate 
the generated candidates and fine-tuned online to capture the 
appearance variations. 
A dynamic switch scheme between local search and image-wide 
re-detection is designed based on a confidence score that is determined by the outputs from both regression and verification networks. 
The experimental results on two recent long-term benchmarks 
demonstrate that our tracker achieves significantly better performance 
than other state-of-the-art algorithms, which can be acted as a new 
baseline for further studies. 

\newpage

{\small
\bibliographystyle{ieee}

}

\end{document}